\documentclass[10pt,twocolumn,letterpaper]{article}

\usepackage{templa}    

\usepackage{algorithm,algorithmicx,setspace}
\usepackage{algpseudocode}

\definecolor{templablue}{rgb}{0.21,0.49,0.74}
\usepackage[pagebackref,breaklinks,colorlinks,allcolors=templablue]{hyperref}

\makeatletter
\def\@makefnmark{}
\makeatother

\makeatletter
\renewcommand{\@fnsymbol}[1]{\ensuremath{\dagger}} 
\makeatother

\title{ActiveInitSplat: How Active Image Selection Helps Gaussian Splatting}

\author{
    \begin{tabular}{c}
        Konstantinos D. Polyzos$^1$*,  Athanasios Bacharis$^2$*, Saketh Madhuvarasu$^1$, \\
        Nikos Papanikolopoulos$^2$, Tara Javidi$^1$ \thanks{The work of Konstantinos D. Polyzos and Tara Javidi was supported by the Eric and Wendy Schmidt AI for Science, and the NSF TILOS AI Institute.} \\ \\
        {\small $^1$ University of California San Diego \; \; $^2$ University of Minnesota} \\
        {\small * Authors declare equal contribution} 
    \end{tabular}
}

\begin{document}
\maketitle

\begin{abstract}
 Gaussian splatting (GS) along with its extensions and variants provides outstanding performance in real-time scene rendering while meeting reduced storage demands and computational efficiency. While the selection of 2D images capturing the scene of interest is crucial for the proper initialization and training of GS, hence markedly affecting the rendering performance, prior works rely on passively and typically densely selected 2D images. In contrast, this paper proposes `ActiveInitSplat', a novel framework for active selection of training images for proper initialization and training of GS. ActiveInitSplat relies on density and occupancy criteria of the resultant 3D scene representation from the selected 2D images, to ensure that the latter are captured from diverse viewpoints leading to better scene coverage and that the initialized Gaussian functions are well aligned with the actual 3D structure. Numerical tests on well-known simulated and real environments demonstrate the merits of ActiveInitSplat resulting in significant GS rendering performance improvement over passive GS baselines in both dense- and sparse-view settings, in the widely adopted LPIPS, SSIM, and PSNR metrics.
\end{abstract}

\section{Introduction}
\label{sec:intro}

3D scene representation and novel view synthesis are fundamental tasks in computer vision that have gained significant traction, driven by their applications in diverse domains such as autonomous driving, robotics, medical imaging, and surveillance, just to name a few. The goal in these tasks is to develop models capable of fast scene rendering while maintaining high visual quality, minimizing storage requirements, and ensuring computational efficiency. Gaussian splatting (GS), first proposed in \cite{kerbl20233d}, along with its extended variants have sparked attention in the research community due to their efficient 3D scene representation via a set of anisotropic Gaussian functions, allowing for fast rendering with memory efficiency, and outstanding visual performance. 

Across different GS models, the selection of initial training images is known to be a crucial factor for the rendering performance, since they have a direct impact on the initialization and training of GS; see e.g. \cite{kerbl20233d}. However, existing GS frameworks hinge on a \textit{passively} pre-selected and typically dense set of training images which ensures redundancy, along with increased processing and computational cost. Even in the few GS frameworks designed for efficient rendering with a limited number of images, passive selection may confine the expressiveness of the 3D scene representation, potentially leading to reduced scene coverage. This paper, in contrast, provides a general framework for \textit{active} and strategic selection of a few yet informative training images captured at diverse viewpoints, to ensure effective GS rendering performance at a lower sampling cost.

\begin{figure*}[!t]
    \centering
    \includegraphics[width=0.9\linewidth]{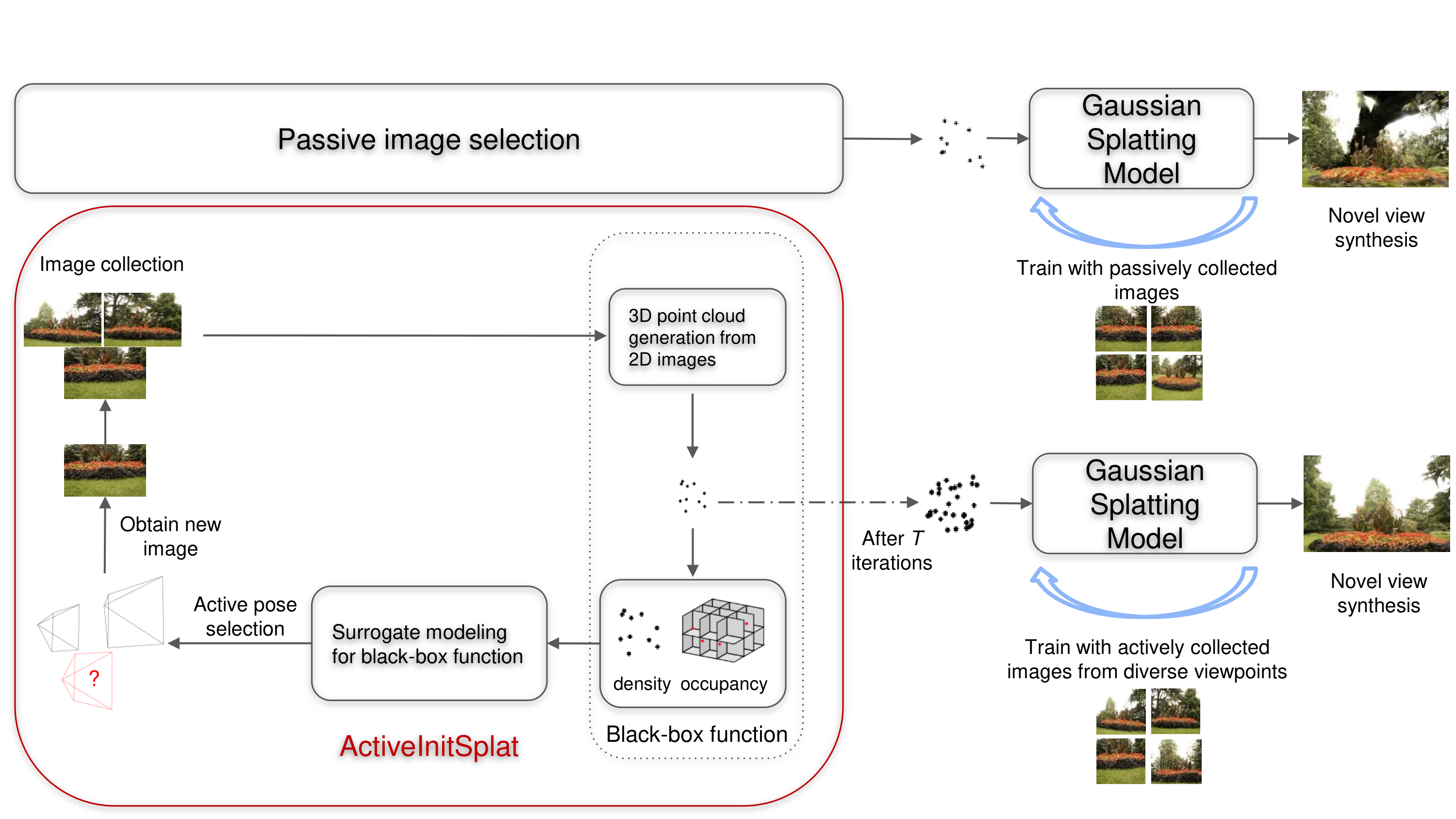}
    \caption{\textbf{ActiveInitSplat in a nutshell}. In contrast to existing Gaussian splatting (GS) methods that rely on passively (and possibly densely) collected 2D images of the scene of interest (upper part), ActiveInitSplat actively selects informative images to assist the initialization and training of GS (lower part). The active selection mechanism lies on optimizing a (black-box) function quantifying the quality of the resultant 3D point cloud from the selected images via density and occupancy criteria. The collected images are captured from diverse viewpoints, ensuring better scene coverage and facilitating the accurate alignment of the initialized Gaussian functions with the underlying 3D structure.}
    \label{fig:ActiveInitSplat}
\end{figure*} 

Although the idea of actively optimizing the training set of images seems natural and necessary, identifying a compact set of optimal camera views that yields high-quality 3D scene representation is fundamentally challenged by the need to guarantee diverse viewpoints without incurring redundancy. Specifically, (i) there is no explicit function in the existing literature that maps a certain set of images to the quality of the reconstructed 3D scene and, consequently, to GS rendering performance; (ii) even if such an explicit function were provided, it certainly would depend on the structure of the scene for which prior information is typically not available; and finally (iii) a sequential processing of multiple different sets of images to empirically assess GS rendering performance is impractical. In the face of these challenges, this paper proposes `ActiveInitSplat', a novel active GS framework shown in Fig. \ref{fig:ActiveInitSplat}, whose contributions can be summarized in the following aspects.
\begin{itemize}
    \item In contrast to existing passive GS approaches, ActiveInitSplat is the first work to introduce an effective and lightweight active camera view selection strategy for Gaussian Splatting without requiring any depth information or any prior knowledge (such as geometric cues) of the 3D scene of interest. 

    \item To guide the active selection process, ActiveInitSplat proposes an innovative optimization formulation, where the objective function integrates density- and occupancy-based criteria of the 3D point cloud, informing the initialization step in GS. As such, our method is agnostic to the subsequent steps of GS. Hence, it can be readily incorporated into the growingly extended variants of GS architectures.



    \item Numerical tests on simulated and well-known real-world benchmark 3D scenes, demonstrate the significance of ActiveInitSplat in actively capturing informative training images from diverse viewpoints for GS, with substantial rendering performance improvement compared to passive selection counterparts.

    \item The proposed active selection mechanism is shown to benefit GS models both in sparse- and dense-view regimes. In addition, ActiveInitSplat performs well both in the general setting of selecting optimal images from any position in 3D space, as well as in scenarios where the view selection is limited to a finite set of images (such as benchmark datasets). 

\end{itemize}

\begin{figure*}[!t]
    \centering
    \includegraphics[width=0.8\linewidth]{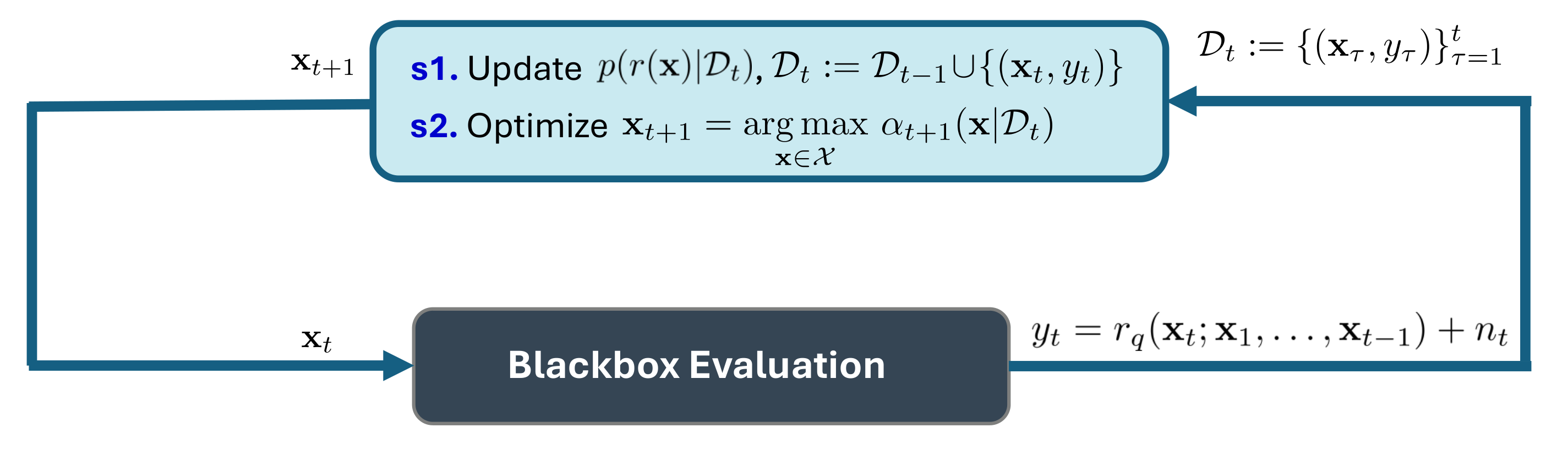}
    \caption{Active viewpoint selection process using Gaussian process-based surrogate modeling for black-box optimization.}
    \label{fig:AL_diagram}
\end{figure*}

\section{Related Works}
\label{sec:related}

In this section, prior art is presented to contextualize the motivation and contributions of ActiveInitSplat. 


\subsection{Novel view synthesis and scene rendering}

\vspace{-0.1cm}

Novel view synthesis (NVS) refers to the problem of estimating views of a 3D scene of interest at unobserved viewpoints/poses based on a certain budget of observed images. NVS initially garnered notable attention with the emergence of approaches leveraging the SfM process, which estimates a (typically sparse) 3D scene representation, and multi-view stereo (MVS) techniques that enable denser 3D reconstructions; see e.g. \cite{kopanas2021point, chaurasia2013depth}. Albeit interesting, the visual quality of the rendered images admitted further improvements. Recently, neural radiance field (NeRF) based methods that leverage multi-layer perceptrons (MLPs) utilizing volumetric ray-marching for NVS, have shown great empirical success. For example, the `Mip-NeRF360' framework in \cite{barron2021mip} has achieved outstanding visual quality of rendered images but may entail up to 48 hours of training, which discourages its application in time-sensitive tasks. To alleviate such computational burdens, existing NeRF-based methods employ simpler MLPs or other techniques; see e.g. \cite{chen2022tensorf, muller2022instant, takikawa2022variable}. Although being more efficient in terms of training, these approaches show inferior rendering quality compared to `Mip-NeRF360'.  

\subsection{Gaussian splatting for novel view synthesis}

Striving for high rendering quality and reduced training time compared to the NeRF-based approaches, the recently developed Gaussian splatting (GS) methods provide principled frameworks to represent a 3D scene of interest using a set of Gaussian functions. To initialize these Gaussian functions so that they are aligned with the 3D structure, the initial GS approach in \cite{kerbl20233d} relies on the SfM process to obtain a representation of the 3D scene. Yet, the SfM process may become inefficient when the number of training images is large, may introduce accumulated errors in the rendering process, and necessitates explicit camera pose knowledge \cite{fan2024instantsplat}. To cope with the latter challenges, alternative pose-free GS approaches have been proposed in \cite{fu2024colmap, jiang2024construct} that rely on monocular depth estimation for 3D representation instead of SfM. However, they may still require extensive training time. Aiming at efficient training solutions, existing SfM-free GS methods \cite{fan2024instantsplat, smart2024splatt3r} capitalize on pre-trained models, including `DUSt3R' \cite{wang2024dust3r} and its following `MASt3R' \cite{leroy2024grounding}, that provide geometry priors for densely sampled, pixel-aligned stereo points. Besides leveraging prior models, alternative SfM-free approaches have been proposed for NVS with efficiency; see e.g. \cite{sun2024correspondence, xu2024freesplatter, zhu2024fsgs, xiong2023sparsegs}. In all these approaches, the selection of the initial set of training images plays a significant role in the rendering performance. Despite their well-documented merits, all these approaches rely on a passively (and possibly densely) pre-selected set of training images,
which may lead to redundant views along with less expressive scene representation and reduced scene coverage, hence motivating the need for active camera view selection to aid GS. Note that the discussion above summarizes key fundamental passive GS methods as representative examples. We omit the many other passive GS variants in the literature, as our main focus is on active GS approaches, which are outlined next.

Lying at the crossroads of robotics and computer vision, existing approaches including \cite{li2024activesplat, jin2024activegs} combine GS models with additional supervision from Voronoi graphs or voxel maps for active path planning, while others leverage GS-based scene representations for active mapping \cite{chen2025activegamer}. Here, we would like to emphasize that these prior works that integrate GS solutions in active mapping or active high-fidelity scene reconstruction for various robotic applications are distinct from the proposed ActiveInitSplat in the sense that the latter aims to find informative camera views for GS beforehand without need for incremental re-training and use of GS. The recent works in \cite{wilson2025pop, xie2025gauss} have explored active view selection in GS through mutual information criteria between residuals and Gaussian parameters or via maximizing P-optimal information gain within a continuously trained GS model. However, these methods require frequent retraining of the GS map as new observations are incorporated or inherently assume an actively maintained GS model, necessitating iterative updates to maintain meaningful uncertainty estimates. In contrast, ActiveInitSplat introduces an active view selection framework that is agnostic to the underlying GS architecture and operates without the need for sequential GS model updates. In robotics settings, the recent work in \cite{strong2024next} seeks instructive camera views for GS accounting for depth uncertainty-based criteria.
Albeit interesting, it is tailored for robotic environments where the targeted scene is mainly focused on specific objects without showing generalizability to broader 3D scenes of interest. In addition, the proposed framework in \cite{strong2024next} integrates ground-truth depth information from depth sensors, which may not be given in several practical settings. Identifying informative camera views for any 3D scene of interest beforehand without the need for re-training of GS during the selection process, and without need for any depth or prior scene information is still unexplored. 

Note that while we acknowledge that part of the existing literature addresses active view selection for NeRF-based methods; see e.g., \cite{pan2022activenerf, xiao2024nerf, kopanas2023improving}, GS models have recently demonstrated state-of-the-art performance in both 3D scene rendering quality and computational efficiency. Therefore, our work, and the related-work discussion, focus exclusively on active view selection for GS models. Notably, since the proposed active view selection strategy in ActiveInitSplat is independent of the subsequent stages of GS models, it can also be readily extended to NeRF-based frameworks.


\section{Preliminaries and Problem Formulation}
\label{sec:preliminaries}

The key concept of GS is to model a 3D scene of interest using a set of $N$ anisotropic Gaussian functions $\{G_i\}_{i=1}^N$ \cite{kerbl20233d}. Relying on an initial set of $T$ pre-selected training images or camera views $\{\mathbf{I}_\tau\}_{\tau=1}^T$ at viewpoints $\{\mathbf{x}_\tau\}_{\tau=1}^T$, conventional GS starts with an (possibly sparse) initial 3D scene representation expressed by a set of 3D points called point cloud (PC), that is typically captured by the SfM process \cite{schoenberger2016sfm}. It is worth noticing that there exist alternative multi-view stereo reconstruction methods for 3D PC prediction that can be used, which aim to overcome the limitations of the traditional SfM process (c.f. Sec. \ref{sec:related}).

For each point $\mathbf{p}_i, i \in \{1,\ldots, N \}$ in the PC, a Gaussian function is defined as 
\begin{align}
G_i(\mathbf{z}) = \alpha_i \text{exp}(-\frac{1}{2}(\mathbf{z}-\mathbf{m}_i)^\top \mathbf{\Sigma}_i^{-1} (\mathbf{z}-\mathbf{m}_i)) \label{eq:Gaus_def}
\end{align}
where $\mathbf{z}$ represents any point in the 3D space that the Gaussian function is evaluated, $\mathbf{m}_i$ is the mean of Gaussian $G_i$ initially centered at the 3D location of $\mathbf{p}_i$, $\mathbf{\Sigma}_i$ is the $3\times 3$ covariance matrix that determines the shape, size and orientation of $G_i$, and $\alpha_i$ is the opacity that indicates the contribution of $G_i$ to the final rendered images. 
Relying on these Gaussian functions, the 2D image rendering formula is \cite{kerbl20233d}
\begin{align}
\mathbf{c}(\mathbf{p}) = \sum_{i=1}^N \mathbf{c}_i G_i^{2D}(\mathbf{p}) \prod_{j=1}^{i-1} \left(1 - G_j^{2D}(\mathbf{p})\right) \label{eq:rend_formula}
\end{align}
where $\mathbf{c}(\mathbf{p})$ represents the color at any pixel $\mathbf{p}$ in the rendered image and $\mathbf{c}_i$ the color corresponding to the $i$th Gaussian that is either expressed via RGB values or spherical harmonics (SH) which are used to model view-dependent color variations. The function $G_i^{2D}(\cdot)$ denotes the projected 2D Gaussian function; see further details e.g. in \cite{kerbl20233d}. 


All Gaussian parameters are trained so that the rendered images match the corresponding training ground-truth ones; i.e. $\{\mathbf{I}_\tau\}_{\tau=1}^T$. To ensure that the covariance matrix $\mathbf{\Sigma}_i$ of each Gaussian $i$ remains positive semi-definite during the optimization process, it is decomposed as $\mathbf{\Sigma}_i = \mathbf{R}_i \mathbf{S}_i \mathbf{S}_i^\top \mathbf{R}_i^\top$ where $\mathbf{S}_i$ is the $3\times3$ diagonal scaling matrix and $\mathbf{R}_i$ represents the rotation matrix analytically expressed in quaternions. The selection of the initial set of camera poses $\{\mathbf{x}_\tau\}_{\tau=1}^T$ that provide the initial training images $\{\mathbf{I}_\tau\}_{\tau=1}^T$, is critical for the GS performance since they affect its initialization and training. Nonetheless, the optimal selection of initial camera poses is a non-trivial task and thus most existing works rely on \textit{passive} selection of training images. In the next section, we will outline our proposed method for \textit{active} selection of proper camera poses $\{\mathbf{x}_\tau^*\}_{\tau=1}^T$ capturing informative camera views $\{\mathbf{I}_\tau^*\}_{\tau=1}^T$ that can assist GS models.


\section{Active Camera View Selection for GS}
\label{sec:method}

Aiming at actively identifying the most informative camera poses $\{\mathbf{x}_\tau^*\}_{\tau=1}^T$ used for proper initialization and training of GS, the present work introduces a novel optimization framework that relies on density and occupancy criteria of the PC in the 3D space emanating from the selected camera views. The intuition is that if the images from the actively selected camera poses result in more dense PCs with better scene coverage compared to passive selection, then (i) the initialized Gaussians will better cover the scene, avoiding missing areas or gaps and (ii) the training images will be selected from diverse viewpoints, thus facilitating the training process of GS. 

At each slot $t$ of the proposed active selection process, the goal is to select the optimal camera pose $\mathbf{x}_t^*$ as 
\begin{align}
    \mathbf{x}_t^* = \underset{\mathbf{x} \in \mathcal{X}}{\arg\max} \; \; r_q(\mathbf{x}; \mathbf{x}_1, \ldots, \mathbf{x}_{t-1})
    \label{eq:opt_goal}
\end{align}
where $\mathcal{X}$ is the feasible set of candidate camera poses and $r_q(\cdot)$ is the objective function that evaluates the quality of the 3D scene representation leveraging density and occupancy-based criteria. Specifically, $r_q(\cdot)$ is expressed as
\begin{align}
    r_q(\mathbf{x}; \mathbf{x}_1, \ldots, \mathbf{x}_{t-1}) := D(\mathcal{P}_C(\mathbf{x}; \mathbf{x}_1, \ldots, \mathbf{x}_{t-1})) \cdot \nonumber \\
    O(\mathcal{P}_C(\mathbf{x}; \mathbf{x}_1, \ldots, \mathbf{x}_{t-1}))  \label{eq:r_opt}
\end{align}
where $\mathcal{P}_C(\mathbf{x}; \mathbf{x}_1, \ldots, \mathbf{x}_{t-1})$ denotes the PC emanating from the images collected from camera poses $\{\mathbf{x}, \mathbf{x}_1, \ldots, \mathbf{x}_{t-1}\}$. The density function $D(\mathcal{P}_C(\mathbf{x}; \mathbf{x}_1, \ldots, \mathbf{x}_{t-1}))$ assesses the density of the PC by measuring the number of 3D points in the PC, and $O(\mathcal{P}_C(\mathbf{x}; \mathbf{x}_1, \ldots, \mathbf{x}_{t-1}))$ quantifies the ``occupancy" of the PC in the 3D space. Inspired by the notion of voxel maps \cite{muglikar2020voxel, hornung2013octomap}, where an environment of interest within a pre-defined bounding box can be represented by a set $\mathcal{V}$ of non-overlapping 3D voxels, the occupancy function $O(\cdot)$ is explicitly defined as
$O(\mathcal{P}_C(\mathbf{x}; \mathbf{x}_1, \ldots, \mathbf{x}_{t-1})) = \sum_{i=1}^{|\mathcal{V}|} v_i/|\mathcal{V}|$,
where $v_i = 1$ only if there is at least one 3D point in $\mathcal{P}_C(\mathbf{x}; \mathbf{x}_1, \ldots, \mathbf{x}_{t-1})$ belonging to voxel $i$, and 0 otherwise.  

The challenge in the optimization problem described in \eqref{eq:opt_goal} is that the objective function $r_q(\cdot)$ cannot be expressed analytically since there is no analytic expression for obtaining $\mathcal{P}_C(\mathbf{x}; \mathbf{x}_1, \ldots, \mathbf{x}_{t-1})$ via SfM or prior models such as MASt3R. In that sense, $r_q(\cdot)$ can be considered as `black-box' function without gradient information, which means that conventional gradient-based solvers may not be applicable. Also note that evaluating $r_q(\mathbf{x}; \mathbf{x}_1, \ldots, \mathbf{x}_{t-1})$ for all $\mathbf{x} \in \mathcal{X}$ may be intractable when the set $\mathcal{X}$ is large. This motivates the use of surrogate modeling to optimize the black-box function $r_q(\cdot)$ as delineated next.  


\subsection{Gaussian process based surrogate model for black-box optimization}

Without analytic expression of $r_q(\cdot)$, a surrogate model can be used to estimate $r_q$ at each slot $t$ using all past observations. Hinging on this surrogate model, an acquisition function (AF) is adopted to find $\mathbf{x}_t^*$ at each slot $t$, which is easy to optimize. While there exist several choices for surrogate modeling, in the present work we will utilize the so-termed Gaussian processes (GPs) that can estimate a nonlinear function along with its probability density function in a data-efficient manner \cite{Rasmussen2006gaussian}. 

Relying on the GP-based surrogate model, the goal at slot $t+1$ is to find $\mathbf{x}_{t+1}^*$ given the training set $\mathcal{D}_t:= \{(\mathbf{x}_\tau, y_\tau)\}_{\tau=1}^t$ consisting of all previous $t$ camera poses collected at matrix $\mathbf{X}_t := \left[\mathbf{x}_1,  \ldots, \mathbf{x}_t\right]^\top$ along with the corresponding (possibly noisy) black-box function evaluations/observations collected at the vector $\mathbf{y}_t := [y_1, \ldots, y_t]^\top$. Here $y_\tau = r_q(\mathbf{x}_\tau; \mathbf{x}_1, \ldots, \mathbf{x}_{\tau-1}) + n_\tau (\forall \tau)$ and $n_\tau \sim \mathcal{N}(0, \sigma_n^2)$ denotes Gaussian noise uncorrelated across $\tau$. Surrogate modeling with GPs deems the sought objective function $r_q(\cdot)$ as random and assumes that it is drawn from a GP prior. This means that the random vector $\mathbf{r}_t:= [r_q(\mathbf{x}_1), r_q(\mathbf{x}_2;\mathbf{x}_1), \ldots, r_q(\mathbf{x}_t; \mathbf{x}_1 \ldots \mathbf{x}_{t-1})]^\top$ containing all function evaluations up until slot $t$, is Gaussian distributed as $p(\mathbf{r}_t| \mathbf{X}_t) = \mathcal{N} ({\bf 0}_t, {\bf K}_t ) (\forall t)$ where $\mathbf{K}_t$ is the covariance matrix whose $(\tau,\tau')$ entry is $[{\bf K}_t]_{\tau,\tau'} = {\rm cov} (r_q(\mathbf{x}_\tau; \mathbf{x}_1, \ldots, \mathbf{x}_{\tau -1}), r_q(\mathbf{x}_\tau'; \mathbf{x}_1, \ldots, \mathbf{x}_{\tau' -1}))$ \cite{Rasmussen2006gaussian}. 

In this setting, the black box function $r_q$ is time-varying since it depends at each slot on all previous camera poses and views collected so far. To capture this dynamic behavior of $r_q$, the covariance between two distinct function evaluations at $\mathbf{x}_\tau$ and $\mathbf{x}_{\tau'}$ is modeled as 
${\rm cov} (r_q(\mathbf{x}_\tau; \mathbf{x}_1, \ldots, \mathbf{x}_{\tau -1}), r_q(\mathbf{x}_\tau'; \mathbf{x}_1, \ldots, \mathbf{x}_{\tau' -1})) = \kappa(\mathbf{x}_\tau, \mathbf{x}_{\tau'}, \tau, \tau')$
where the kernel function $\kappa(\mathbf{x}_\tau, \mathbf{x}_{\tau'}, \tau, \tau')$ is decomposed as 
\begin{align}
\kappa(\mathbf{x}_\tau, \mathbf{x}_{\tau'}, \tau, \tau') =  \kappa_s (\mathbf{x}_\tau, \mathbf{x}_{\tau'}) \cdot \kappa_{\text{time}}(\tau, \tau') 
\label{eq:kernel_temp}
\end{align}
with $\kappa_s (\mathbf{x}_\tau, \mathbf{x}_{\tau'})$ measuring the pairwise similarity between the camera poses $\mathbf{x}_\tau$ and $\mathbf{x}_{\tau'}$, and $\kappa_{\text{time}}(\tau, \tau')$ quantifying the distance between slots $\tau$ and $\tau'$ with time-decaying behavior. This means that $\kappa_{\text{time}}(\tau, \tau')$ becomes smaller as the distance between $\tau$ and $\tau'$ becomes larger, which implies that the function evaluations $r_q(\mathbf{x}_\tau; \mathbf{x}_1, \ldots, \mathbf{x}_{\tau -1}), r_q(\mathbf{x}_\tau'; \mathbf{x}_1, \ldots, \mathbf{x}_{\tau' -1})$ are highly correlated when $\tau$ is not far away from $\tau'$. The intuition is that the selection of camera poses and corresponding views should depend more on the most recent observations compared to the oldest ones since the most recent observations carry the information of more collected camera views.   

With the GP prior at hand, it can be shown that the posterior pdf of $r_q$ for any (unseen) view $\mathbf{x} \in \mathcal{X}$ at slot $t+1$ is Gaussian distributed as  $p(r_q({\bf x})|\mathcal{D}_{t}) = \mathcal{N}(\mu_t ({\bf x}), \sigma_{t}^2 ({\bf x}))$ with mean and variance expressed analytically as \cite{Rasmussen2006gaussian}
\begin{subequations}
\begin{align}	
\mu_{t+1} ({\bf x}) &= \mathbf{k}_{t+1}^{\top} ({\bf x}) (\mathbf{K}_t + \sigma_{n}^2
 \mathbf{I}_t)^{-1} \mathbf{y}_t \label{eq:mean}\\
\sigma_{t+1}^2 ({\bf x}) &= \!\kappa(\mathbf{x},\mathbf{x}, t+1, t+1)\!  \nonumber \\ &\quad
- \mathbf{k}_{t+1}^{\top} ({\bf x}) (\mathbf{K}_t\! +\! \sigma_{n}^2 \mathbf{I}_t)^{-1} \mathbf{k}_{t+1} ({\bf x}) \label{eq:variance}
\end{align}\label{eq:plain_gpp}
\end{subequations}
where $\mathbf{k}_{t+1} ({\bf x}) := [\kappa(\mathbf{x}_1, \mathbf{x}, 1, t+1), \ldots, \kappa(\mathbf{x}_t,  \mathbf{x}, t, t+1)]^\top$. The mean in \eqref{eq:mean} provides a point estimate of the function evaluation at $\mathbf{x}$ and the variance in \eqref{eq:variance} quantifies the associated uncertainty. Relying on the surrogate model $p(r_q({\bf x})|\mathcal{D}_{t})$, the next optimal camera pose $\mathbf{x}_{t+1}^*$ is identified optimizing the AF $\alpha(\mathbf{x}|\mathcal{D}_t)$. Here, the camera pose at slot $t+1$ is selected solving the optimization problem 
\begin{align}
    \mathbf{x}_{t+1} = \underset{\mathbf{x}\in \mathcal{X}}{\arg\max} \; \alpha(\mathbf{x}|\mathcal{D}_t) := \mu_{t+1}(\mathbf{x}) 
    \label{eq:acquisition}
\end{align}
where it is intuitive that if the surrogate model captures well the objective function, then selecting $\mathbf{x}$ that maximizes the posterior mean in \eqref{eq:mean} most likely provides the maximum of $r_q$ at this slot. Upon selecting $\mathbf{x}_{t+1}$, the corresponding (possibly noisy) function evaluation $y_{t+1}$ is obtained, and the training set for the surrogate model is augmented as $\mathcal{D}_{t+1} = \mathcal{D}_t \bigcup \{(\mathbf{x}_{t+1}, y_{t+1})\}$ for the next iteration. This process is repeated up until slot $T$, as shown in Fig. \ref{fig:AL_diagram}. Then, the informative set of the $T$ collected camera poses and views can be used for initialization and training of different GS models. The steps of ActiveInitSplat are outlined in Algorithm \ref{Alg:ActiveInitSplat}.

\vspace{0.1cm}

\noindent \textbf{Remark 1.} To the best of our knowledge, ActiveInitSplat is the first framework to introduce active view selection based on the 3D representation quality derived from the selected images, to facilitate GS initialization and training while remaining agnostic to subsequent GS steps. In this work, 3D representation quality is evaluated using density and occupancy criteria of the 3D PC; however, we would like to note that incorporating additional metrics could further enhance ActiveInitSplat’s performance. Potential extensions include connectivity metrics to detect gaps or discontinuities in otherwise continuous structures, and sharpness metrics to assess edge and detail clarity, among others. Integrating such criteria into the active selection process constitutes part of our future research agenda.

\begin{algorithm}[t]
\caption{ActiveInitSplat algorithm.} 
\label{Alg:ActiveInitSplat}
\begin{algorithmic}[1]
\State  \textbf{Initialization:} Images $\{\mathbf{I}_{0}, \mathbf{I}_1\}$, camera poses $\{\mathbf{x}_{0},\mathbf{x}_1\}$ 
\State Set $y_{0}=0$
\State Compute $y_1 = r_q(\mathbf{x}_1; \mathbf{x}_0) + n_1$ via \eqref{eq:r_opt}. 
\State Construct initial (training) set $\mathcal{D}_1 = \{(\mathbf{x}_i, y_i)\}_{i=0}^1$
\State \# Active camera view selection process

\For{$t = 2, \ldots, T$}

\State Update parameters of kernel $\kappa$ maximizing the marginal likelihood. 

\State Update GP surrogate model $p(r_q({\bf x})|\mathcal{D}_{t}) = \mathcal{N}(\mu_t ({\bf x}), \sigma_{t}^2 ({\bf x}))$ via \eqref{eq:mean} and \eqref{eq:variance}.

\State Obtain next camera pose $\mathbf{x}_{t+1}$ via \eqref{eq:acquisition} and obtain corresponding image $\mathbf{I}_{t+1}$.

\State Compute $y_{t+1} = r_q(\mathbf{x}_{t+1}; \mathbf{x}_0,\ldots, \mathbf{x}_t) + n_{t+1}$ via \eqref{eq:r_opt}. 

\State Augment training set $\mathcal{D}_{t+1} = \mathcal{D}_t \bigcup \{(\mathbf{x}_{t+1}, y_{t+1})\}$.

\EndFor

\State \# Gaussian splatting initialization and training

\State Initialize Gaussian functions $\{G_i\}_{i=1}^N$ using $\mathcal{P}_C(\mathbf{x}_{T}; \mathbf{x}_0, \ldots, \mathbf{x}_{T-1})$. 
\State Train a certain GS model using $\{\mathbf{I}_0,\ldots,\mathbf{I}_{T}\}$.

\end{algorithmic}
\end{algorithm}


\begin{table*}[t]
\begin{center}
\scalebox{0.88}{
    \begin{tabular}{lccccccc}
    \toprule
    \textbf{Scene} & \textbf{Method} &  $r_q$ \textbf{value}$\uparrow$ & \textbf{LPIPS}$\downarrow$ & \textbf{SSIM}$\uparrow$ & \textbf{PSNR}$\uparrow$ \hspace{-2mm} \\
    \midrule
    &3DGS-passive-random  &  25.66652 & 0.3565 & 0.5672 & 18.796 \\
    bonsai &3DGS-passive-standard  & 27.83765 & 0.3108 & 0.6089 & 18.808  \\
    &ActiveInitSplat  & \textbf{35.43626} & \textbf{0.2563} & \textbf{0.7221} & \textbf{22.602}  \\
    \hline
    &3DGS-passive-random  &  1.05133 & 0.4619 & 0.3104 & 14.628   \\
    flowers&3DGS-passive-standard  & 1.51089 & 0.4606 & 0.3084 & 15.318   \\
    &ActiveInitSplat  & \textbf{2.58819} & \textbf{0.4025} & \textbf{0.4191} & \textbf{17.538}   \\
    \hline
    &3DGS-passive-random  & 0.4453 & 0.4402 & 0.3775 & 15.714  \\
    bicycle&3DGS-passive-standard  & 0.79184 & 0.4313 & 0.3841 & 17.176  \\
    &ActiveInitSplat  & \textbf{0.97178} & \textbf{0.4287} & \textbf{0.4157} & \textbf{17.871}  \\
    \hline
    & 3DGS-passive-random  & 0.14839 & 0.3669 & 0.5392 & 16.847  \\
    treehill&3DGS-passive-standard  & 0.15071 & 0.3699 & 0.5454 & 17.077 \\
    &ActiveInitSplat  & \textbf{0.17711} & \textbf{0.3649} & \textbf{0.5893} & \textbf{18.023}  \\
    \hline
    &3DGS-passive-random  & 20.36397 & 0.1699 & 0.7446 & 24.031  \\
    garden&3DGS-passive-standard  & 26.13629 & 0.2070 & 0.6587 & 22.772 \\
    &ActiveInitSplat  & \textbf{26.9456} & \textbf{0.1678} & \textbf{0.7536} & \textbf{24.384}  \\
    \hline
    &3DGS-passive-random  & 0.00314 & 0.49 & 0.756 & 18.88   \\
    corn&3DGS-passive-standard  & 0.00404 & 0.456 & 0.77 & 20.3  \\
    &ActiveInitSplat  & \textbf{0.00409} & \textbf{0.431} & \textbf{0.792} &  \textbf{21.649} \\
    \hline
    \end{tabular}}
\caption{$r_q$, LPIPS, SSIM and PSNR values for all competing methods on real-world 3D scenes using COLMAP for PC generation in the view selection process for GS.}
\label{tab:quant}
\end{center}
\vspace{-4mm}
\end{table*}


\begin{figure*}[!t]
    \centering
    \includegraphics[width=0.88\linewidth]{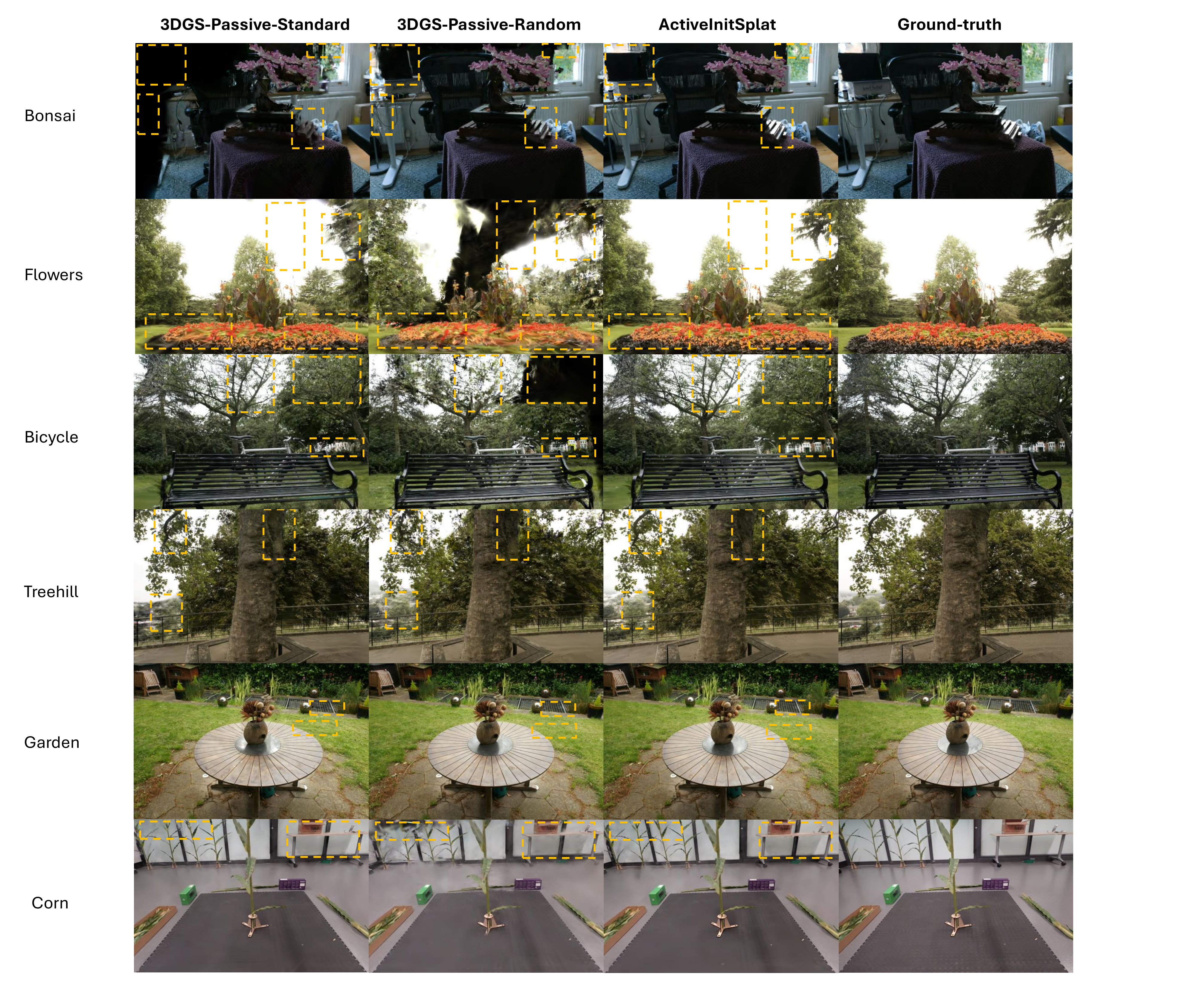}
    \caption{Visual comparison of ActiveInitSplat with the passive selection counterparts in real-world datasets. For each dataset, we illustrate a single indicative test image where the differences are shown with the annotated yellow dashed boxes. When using ActiveInitSplat, the rendered images are closer to the ground-truth ones.}
    \label{fig:qual_plot}
\end{figure*}

\section{Numerical Evaluation}
\label{sec:numerical_tests}

This section demonstrates the merits of the advocated ActiveInitSplat with conducted numerical tests on simulated and real-world datasets. 

\subsection{Implementation details}
\label{subsec:implem}

The rendering performance of ActiveInitSplat for NVS is initially assessed on real-world datasets including a suite of 3D scenes from the database in \cite{barron2022mipnerf360} and the corn dataset representing a scene of a corn plant inside a lab. 
Specifically, the `bonsai', `flowers', `bicycle', `treehill', and 'garden' from \cite{barron2022mipnerf360} comprise in total $M=292, 173, 194, 141, 185$ images respectively, and the `corn' dataset consists of $M=77$ images. To evaluate the generalization rendering performance on unseen novel views, $N_{\text{test}} = 20$ out of the total $M$ images are used for testing, and in our experimental setting the remaining $M-N_{\text{test}}$ images are considered as candidate views, from which $T<M-N_{\text{test}}$ are actively selected from ActiveInitSplat. For the bonsai, flowers, bicycle, and garden datasets ActiveInitSplat is run for $T=70$ iterations, whereas for the treehill and corn datasets we consider $T=60$ and $T=30$ respectively, since the latter datasets comprise in total a smaller number of images compared to the former ones.

In these datasets, we assume access to the poses of the candidate views, whose estimate is already provided by COLMAP \cite{schoenberger2016sfm} in the database \cite{barron2022mipnerf360}; though other pose-estimation techniques can be utilized such as \cite{leroy2024grounding}. It is worth noticing that in practical settings, ActiveInitSplat does not require to be confined to a finite set of candidate views. Specifically, it can identify the most informative ground-truth pose $\mathbf{x}_t$ at each slot $t$ of the active selection process, searching at the entire 3D space that the scene of interest belongs to, instead of being confined to a (possibly limited) set of candidate views. In the next subsection, we will additionally present a realistic ablation study on a simulated environment where ActiveInitSplat can select at each iteration any camera view in the 3D space. For proper evaluation, ActiveInitSplat is tested for five different test sets of $N_{\text{test}}=20$ independently. In each independent run, $T$ views are actively selected from the remaining $M-N_{\text{test}}$ candidate views. The final evaluation considers the average rendering performance across all five runs. The latter is quantified using the LPIPS, SSIM, and PSNR metrics that are widely adopted for NVS performance assessment. The active selection process of ActiveInitSplat is compared against two passive view selection approaches used for GS; that is (i) `3DGS-passive-standard' that selects views sequentially in the standard way each dataset is constructed (typically following a `circular' formation of collecting images), and (ii) `3DGS-passive-random' that randomly selects a view from the (remaining) candidate views at each iteration. To avoid overstating the benefits of the advocated active selection process, `3DGS-Passive-random' considers the best performing of five independent runs. The training of 3DGS relies on the L1 and D-SSIM loss function as in \cite{kerbl20233d}, and is carried out using 30000 iterations for all datasets except the corn that is trained for 10000 iterations. 

For demonstration purposes, the 3D representation $\mathcal{P}_C(\mathbf{x}_t; \mathbf{x}_1, \ldots, \mathbf{x}_{t-1})$ used for the black-box function evaluation $r_q(\mathbf{x}_t; \mathbf{x}_1 \ldots \mathbf{x}_{t-1})$ upon acquiring $\mathbf{x}_t$ at slot $t$, is estimated using COLMAP \cite{schoenberger2016sfm} similarly as in the traditional SfM-based GS model in \cite{kerbl20233d}. 
However, we note that other alternative techniques for obtaining $\mathcal{P}_C(\cdot)$ from 2D images can be readily integrated into our ActiveInitSplat framework; see e.g. a related ablation study in \ref{sec:as1}. For the Gaussian process surrogate model, the kernel functions $\kappa_s, \kappa_t$ used for computing $\kappa$ in \eqref{eq:kernel_temp} are selected to be RBF kernels, which is a typical choice across different GP-based tasks \cite{Rasmussen2006gaussian}. The kernel hyperparameters are estimated maximizing the marginal likelihood using the \textit{sklearn} package and are re-fitted at the end of each iteration where a newly acquired $(\mathbf{x}_t, y_t)$ pair of camera-view and (noisy) function evaluation becomes available. Note that the  kernel function has a significant role in the GP surrogate model and existing works aim to learn the proper kernel form; see e.g. \cite{malkomes2016bayesian, teng2020scalable, lu2023surrogate, duvenaud2013structure}, that could be applied in our ActiveInitSplat but exceeds the scope of the current work.   

The active selection process (lines 6–12 of Algorithm \ref{Alg:ActiveInitSplat}) was executed on an Intel Core i7-5930K CPU and a GeForce Tatan X GPU, while the GS training process (lines 14–15) was conducted on a  Tesla V100-SXM2-16GB GPU hosted on Amazon servers. In experiments utilizing the SfM process for PC generation, the COLMAP software was employed, while for those leveraging the MASt3R pre-trained model in subsection \ref{sec:as1}, we used the implementation available at \url{https://github.com/naver/mast3r}. For the GS training and rendering process, we relied on the implementation from \url{https://github.com/graphdeco-inria/gaussian-splatting}.

\begin{figure*}[!t]
    \centering
    \includegraphics[width=0.95\linewidth]{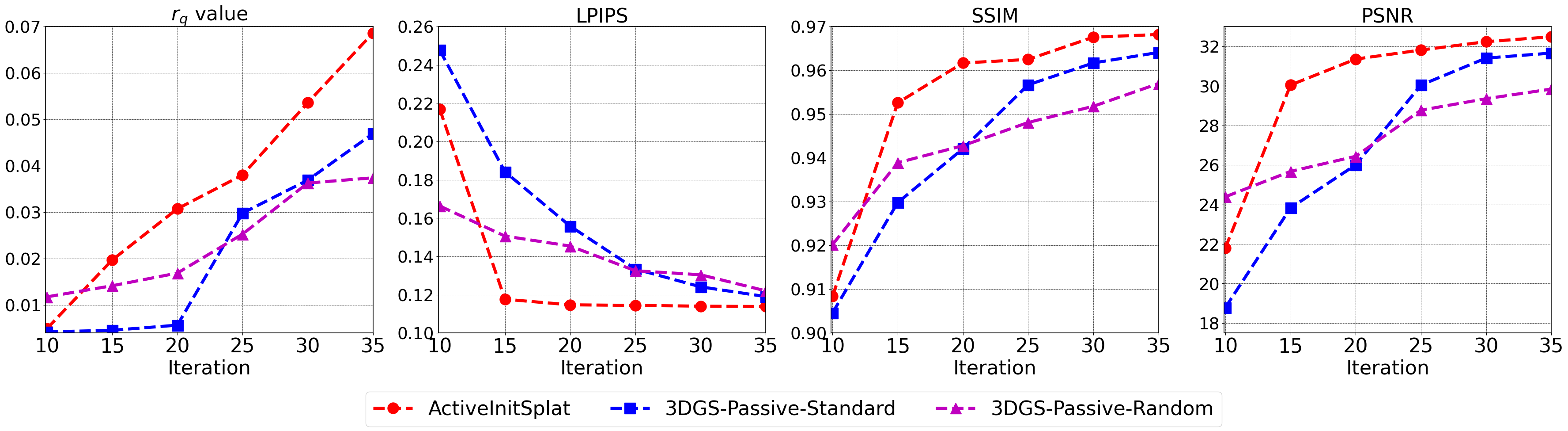}
    \caption{The value of (a) $r_q$ function, (b) LPIPS, (c) SSIM, and (d) PSNR at each iteration of the selection process for all competing methods in the simulated `office 4' environment \cite{replica19arxiv} where any viewpoint in the 3D space can be selected at each iteration. After 10 iterations, ActiveInitSplat results in higher quality 3D scene representation compared to baselines, along with superior GS rendering performance. In addition, ActiveInitSplat converges faster to the desired rendering performance. }
    \label{fig:sim_plot}
\end{figure*}

\begin{table*}[t]
\begin{center}
\scalebox{0.88}{  
    \begin{tabular}{lccccccc}
    \toprule
    \textbf{Scene} & \textbf{Method} &  $r_q$ \textbf{value}$\uparrow$ & \textbf{LPIPS}$\downarrow$ & \textbf{SSIM}$\uparrow$ & \textbf{PSNR}$\uparrow$ \hspace{-2mm} \\
    \midrule
    &3DGS-passive-random  & 26.913 & 0.4652& 0.5829 & 15.796 \\
    bonsai &3DGS-passive-standard  & 105.711 & 0.4231 & 0.5562 & 16.033 \\
    &ActiveInitSplat  & \textbf{173.865} & \textbf{0.4120} & \textbf{0.6041} & \textbf{18.121} \\
    \hline
    &3DGS-passive-random  & 29.295 & 0.5677  & 0.3606 & 17.708  \\
    garden&3DGS-passive-standard  & 27.851 & 0.5518  & 0.3382 & 17.654 \\
    &ActiveInitSplat  & \textbf{47.723} & \textbf{0.5348}  & \textbf{0.3878} & \textbf{17.806}   \\
    \hline
    &3DGS-passive-random  & 1.368& 0.4940  & 0.7770 & 20.203   \\
    corn&3DGS-passive-standard  & 1.392 & 0.4942  & 0.7787 & 20.452  \\
    &ActiveInitSplat  & \textbf{1.58} & \textbf{0.4916}  & \textbf{0.7811} & \textbf{20.584}  \\
    \hline
    \end{tabular}}
\caption{$r_q$, LPIPS, SSIM, and PSNR values for all competing methods on three indicative real-world 3D scenes using the MASt3R model for PC generation within the view selection process for GS in sparse-view settings.}
\label{tab:quant_mast3r}
\end{center}
\vspace{-4mm}
\end{table*}


\begin{figure*}[!t]
    \centering
    \includegraphics[width=0.999\linewidth]{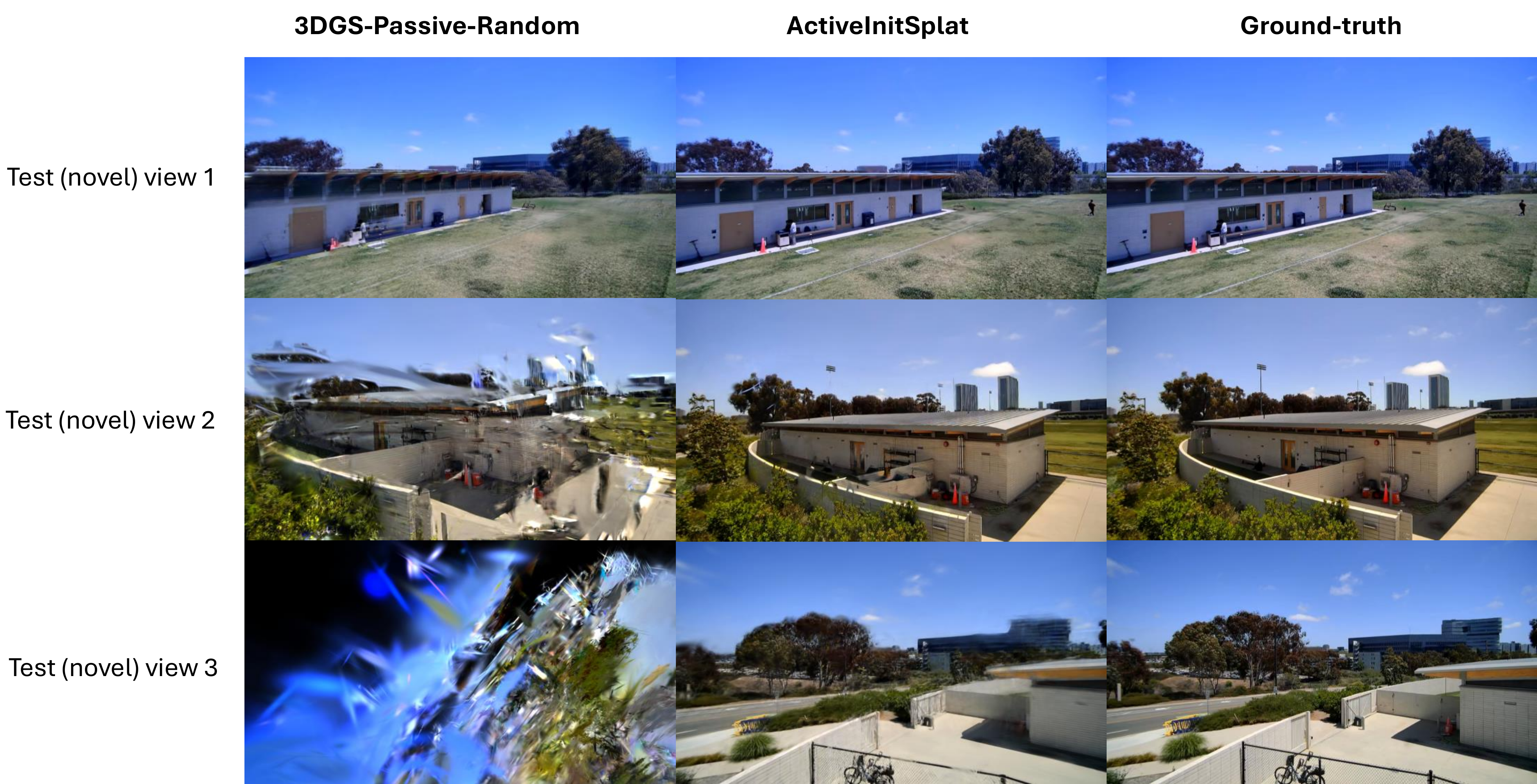}
    \caption{Visual comparison of ActiveInitSplat with the passive selection counterparts in the real-world drone-test platform on three indicative test (novel) viewpoints. These test views are captured from diverse angles and include not only the building within the sports field but also the broader surroundings, including background buildings, trees, and roads, which are particularly challenging to capture in such a large-scale environment. It is evident that the rendered images from ActiveInitSplat have substantially improved quality compared to the passive counterpart.}
    \label{fig:real_qual_plot}
\end{figure*}

\subsection{Numerical results}

To showcase the merits of ActiveInitSplat, we provide numerical results on widely used synthetic and real benchmark datasets, as well as on a real-world drone testing platform, supplemented by additional ablation studies.
 
\subsubsection{Numerical tests on well-known benchmark datasets}

To assess the performance of ActiveInitSplat in the 3D scenes described in subsection \ref{subsec:implem}, the first step is to corroborate the effectiveness of ActiveInitSplat in identifying camera views from diverse viewpoints that provide high-quality 3D scene representation with increased scene coverage. To that end, Table \ref{tab:quant} initially demonstrates the $r_q$ function value of all competing methods, that quantifies the quality of a point cloud in terms of density and scene coverage given a specific set of camera views (c.f. section \ref{sec:method}). It can be seen in Table \ref{tab:quant} that ActiveInitSplat attains a higher $r_q$ value compared to the passive selection counterparts in all datasets illustrating the effectiveness of the proposed GP-based black-box optimization technique in actively capturing informative camera views. With the selection of informative 2D images granted, the goal next is to evaluate how this active selection translates into the GS rendering performance for NVS. Table \ref{tab:quant} additionally reports the LPIPS, SSIM, and PSNR rendering metrics for all different methods upon completing $T$ iterations of the selection process. It is evident that ActiveInitSplat is the best-performing approach in all datasets and in all rendering metrics implying that the passive methods require a denser selection of images to meet the rendering performance of ActiveInitSplat. Besides quantitative evaluations, Fig. \ref{fig:qual_plot} also shows qualitative results of indicative rendered test images compared to the corresponding ground-truth ones in all tested 3D scenes. The annotated dashed yellow boxes in Fig. \ref{fig:qual_plot} indicate the differences in the rendered images between all competing methods where it is shown that the rendered images from ActiveInitSplat are closer to the ground-truth ones compared to the passive baselines. 

\begin{table}[t]
    \centering
    \caption{LPIPS, SSIM and PSNR values for competing methods in the real-world drone-test platform.}
    \scalebox{0.88}{
    \begin{tabular}{lccc}
        \toprule
        \textbf{Method} & \textbf{LPIPS} $\downarrow$ & \textbf{SSIM} $\uparrow$ & \textbf{PSNR} $\uparrow$ \\
        \midrule
        3DGS-passive-random    & 0.4727 & 0.4566 & 13.893 \\
  ActiveInitSplat        & \textbf{0.3273} & \textbf{0.7024} & \textbf{20.686} \\
        \bottomrule
    \end{tabular}}
    \label{tab:drone_test_platform}
\end{table}

\begin{table*}[t]
\begin{center}
\scalebox{0.88}{
\begin{tabular}{lcccc}
\toprule
\textbf{Scene} & \textbf{ActiveInitSplat (T=12)}  & \textbf{GS-passive-random (T=12)}  & \textbf{GS-passive-random (T=20)} & \textbf{GS-passive-random (T=30)} \\
\midrule
Bonsai  & 5.78 mins & 5.22 mins & 26.88 mins & 164.18 mins \\
Corn    & 7.13 mins & 6.76 mins & 29.31 mins & 164.51 mins \\
Garden  & 8.48 mins & 8.15 mins & 62.03 mins & 211.36 mins \\
\bottomrule
\end{tabular}}
\caption{Total runtime comparison between ActiveInitSplat and passive GS alternatives for collecting and processing $T$ camera views for 3D PC generation used for GS initialization.}
\label{tab:running_time}
\end{center}
\end{table*}

\begin{table*}[t]
\begin{center}
\scalebox{0.88}{
    \begin{tabular}{lccccccc}
    \toprule
    \textbf{Scene} & \textbf{Method} &  $r_q$ \textbf{value}$\uparrow$ & \textbf{LPIPS}$\downarrow$ & \textbf{SSIM}$\uparrow$ & \textbf{PSNR}$\uparrow$ \hspace{-2mm} \\
    \midrule
    &CompactGS-passive-random  &  25.66652 & 0.3612 & 0.5669 & 18.8106 \\
    &CompactGS-passive-standard  & 27.83765 & 0.3164 & 0.5929 & 18.7264  \\
    &ActiveInitSplat \& CompactGS  & \textbf{35.43626}  & \textbf{0.2649} & \textbf{0.7164} & \textbf{22.5226}   \\
    bonsai&CoRGS-passive-random  &  25.66652 & 0.3604 & 0.5738 & 18.9091 \\
    &CoRGS-passive-standard  & 27.83765 & 0.3483 & 0.5808 & 17.9609  \\
    &ActiveInitSplat \& CoRGS  & \textbf{35.43626} & \textbf{0.2891}  & \textbf{0.7046} & \textbf{21.3254}   \\
    \hline
    & CompactGS-passive-random  & 0.14839 & 0.3869 & 0.5441 & 17.0439  \\
    &CompactGS-passive-standard  & 0.15071 & 0.3834 & 0.5513 & 17.2222 \\
    &ActiveInitSplat \& CompactGS  & \textbf{0.17711} & \textbf{0.3816} & \textbf{0.5902} & \textbf{17.9837}  \\
    treehill & CoRGS-passive-random  & 0.14839 & \textbf{0.3497}  & 0.5711 & 18.9142 \\
    &CoRGS-passive-standard  & 0.15071 & 0.3748 & 0.5542 & 18.6461 \\
    &ActiveInitSplat \& CoRGS  & \textbf{0.17711} & 0.3665  & \textbf{0.6023} & \textbf{19.9442} \\
    \hline
    &CompactGS-passive-random  & 0.00314 & 0.4985 & 0.7491 & 18.7709   \\
    &CompactGS-passive-standard  & 0.00404 & 0.4613 & 0.7691 & 20.2001  \\
    &ActiveInitSplat \& CompactGS  & \textbf{0.00409} & \textbf{0.4361}  & \textbf{0.7934} & \textbf{21.4923} \\
    corn&CoRGS-passive-random  & 0.00314 & 0.4981 & 0.7505 & 18.8101   \\
    &CoRGS-passive-standard  & 0.00404 & 0.4722 & 0.7702 & 20.2098  \\
    &ActiveInitSplat \& CoRGS  & \textbf{0.00409} & \textbf{0.4402}  & \textbf{0.7983} & \textbf{21.6326}   \\
    \hline
    \end{tabular}}
\caption{$r_q$, LPIPS, SSIM and PSNR values for all competing methods on real-world 3D scenes using CompactGS and CoRGS architectures and COLMAP for PC generation.}
\label{tab:corcompact}
\end{center}
\vspace{-4mm}
\end{table*}

\subsubsection{Numerical tests on a real-world drone-test platform}

Having demonstrated the benefits of ActiveInitSplat over passive GS counterparts on widely used benchmark datasets with relatively small scenes, in this section we aim to evaluate its rendering performance in a real-world experiment using a drone-test platform over a large scene centered on a building within a sports field. Note that, unlike the benchmark datasets where images are typically captured by humans using cell phones around the scene of interest, capturing comprehensive views in the large-scale drone test platform is impractical when using human cell phones, and constitutes a non-trivial task. In addition to the inherent challenges posed by large-scale environments, 3D rendering on the real-world drone test platform is markedly challenged by: (i) the lower resolution of the extracted images compared to those in the benchmark datasets used in Section 5 of the main paper; (ii) noise in the extracted images introduced by drone movement and adverse weather conditions, such as the strong winds that were encountered while conducting the experiment; and (iii) scene dynamics, including moving vehicles and train, along with tree/plant movements due to wind effects. To that end, this experiment can further highlight the practical utility of active image selection for Gaussian Splatting in more complex, large-scale and challenging real-world environments.

For this experiment, we used a Holybro X500 drone equipped with an SIYI A8 camera to extract images. For drone/camera localization, the reference origin used to obtain the drone position is placed in front of the building. Utilizing this reference origin, the occupancy map/ bounding box required to compute the occupancy score in \eqref{eq:r_opt} is constructed with boundaries defined by the outer corners of the sports field. Similarly as in Section 5 of the main paper, for the 3DGS-passive-random baseline, we consider the best-performing in terms of rendering quality out of five independent runs. Note that the `3DGS-passive-standard' baseline is omitted here, as there is no pre-constructed dataset following a specific image acquisition pattern, as in the benchmark datasets. Moreover, adopting a standard circular image collection strategy to form a `3DGS-passive-standard' baseline would certainly require a significantly larger number of images in this large-scale environment, leading to increased computational overhead. 
For the image collection process used for GS initialization and training, we run both ActiveInitSplat and the passive GS baselines for $T=100$ iterations.

To assess the performance of all competing methods on previously unseen novel (test) views, we illustrate in Fig. 5 the rendered images from ActiveInitSplat and passive GS baseline on three indicative testing viewpoints compared to the ground-truth testing (novel) views. Note that these representative test views are selected from diverse angles and viewpoints, capturing various parts of the scene, including the building within the sports field as well as the broader surroundings such as background buildings, trees, and roads, which are particularly challenging to render accurately. As it can be clearly seen in Fig. 5, the rendered image produced by ActiveInitSplat for the first test viewpoint more closely resembles the ground truth compared to the `3DGS-passive-random' baseline, while in the other two test viewpoints ActiveInitSplat exhibits substantially improved rendering quality performance. Notably, the challening third test view, which predominantly covers regions outside the sports field, is effectively captured by ActiveInitSplat, in contrast to the passive baseline which demonstrates poor rendering performance. This highlights the effectiveness of ActiveInitSplat in selecting informative images with broad scene coverage, facilitating efficient GS initialization and training, and enabling strong generalization across diverse test viewpoints. Lastly, Table \ref{tab:drone_test_platform} reports the LPIPS, SSIM, and PSNR quantitative metrics, showing that ActiveInitSplat achieves substantially superior rendering performance compared to the passive GS baseline. This indicates that the passive approach would require a much denser image collection and processing, and consequently greater computational overhead, to match the performance of ActiveInitSplat.

\subsubsection{Additional ablation studies}

Next, we pose and aim to answer the following research questions to further highlight the merits of ActiveInitSplat. 

\vspace{0.1cm}

\noindent \textit{\textbf{Q1.} How does ActiveInitSplat perform in realistic settings with no constraint on candidate views, i.e. where any viewpoint in the 3D space of interest can be selected?}\label{sec:as1}

\vspace{0.1cm}

\noindent To assess ActiveInitSplat in realistic settings where any viewpoint in the 3D space of interest can be selected, we leverage the simulated 3D environment representing the `office 4' scene from \cite{replica19arxiv}, where images are rendered using the PyRender software \cite{pyrender}. In this setting, $N_{\text{test}} = 50$ test views are considered to evaluate the generalization rendering performance on unseen novel views. At each iteration $t$ of the view selection process, the proposed approach can actively select any camera pose $\mathbf{x}_t$ in the 3D space of the simulated environment and collect the corresponding RGB view $\mathbf{I}_t$. For the `3DGS\_Passive\_Standard' method, images are selected incrementally following a circular path at a fixed height. Note that in this ablation study we also aim to show that ActiveInitSplat converges faster to the desired rendering performance compared to the passive counterparts. To that end, Fig. \ref{fig:sim_plot} illustrates the $r_q$ function value along with the LPIPS, SSIM, and PSNR values at every 5 iterations of the camera view selection process for all competing methods. It can be observed that after 10 iterations, ActiveInitSplat enjoys higher $r_q$ function score compared to the passive baselines, leading to higher-quality 3D representations from a more diverse set of camera views. It is worth mentioning that `3DGS-Passive-Random' outperforms ActiveInitSplat in the 10th iteration because its reported result corresponds to the best-performing run among five independent trials, increasing the likelihood of capturing a more informative set of images when only ten are collected. Regarding rendering performance on test views, Fig. \ref{fig:sim_plot} clearly shows that after 10 iterations, ActiveInitSplat outperforms all other approaches across all iterations for the LPIPS, SSIM, and PSNR metrics, as anticipated from its higher $r_q$ values. It is also evident that ActiveInitSplat has much faster convergence to the best value compared to the passive selection baselines for all three metrics. For instance, in the LPIPS metric, ActiveInitSplat reaches a lower LPIPS value by iteration 15, whereas both passive baselines require 35 iterations to achieve a comparable result.

\begin{table}[t]
\centering
\caption{Comparison between ActiveInitSplat and active coverage-based 3DGS-FVS baseline on multiple 3D scenes based on the LPIPS, SSIM, and PSNR metrics.}
\label{tab:fvs_comparison}
\scalebox{0.88}{
\begin{tabular}{l l c c c}
\toprule
\textbf{Scene} & \textbf{Method} & \textbf{LPIPS} ↓ & \textbf{SSIM} ↑ & \textbf{PSNR} ↑ \\
\midrule
Flowers  & ActiveInitSplat & \textbf{0.4025} & \textbf{0.4191} & \textbf{17.538} \\
         & 3DGS-FVS        & 0.4317 & 0.3296 & 16.455 \\
\hline
Bicycle  & ActiveInitSplat & 0.4287 & 0.4157 & \textbf{17.871} \\
         & 3DGS-FVS        & \textbf{0.4196} & \textbf{0.4239} & 17.551 \\
\hline
Treehill & ActiveInitSplat & \textbf{0.3649} & \textbf{0.5893} & \textbf{18.023} \\
         & 3DGS-FVS        & 0.3759 & 0.5569 & 16.981 \\
\hline
Garden  & ActiveInitSplat  & \textbf{0.1678} & \textbf{0.7536} & \textbf{24.384} \\
        & 3DGS-FVS         & 0.1745 & 0.7526 & 24.162 \\
\hline
Bonsai  & ActiveInitSplat  & 0.2563 & 0.7221 & 22.602 \\
        & 3DGS-FVS         & \textbf{0.2303} & \textbf{0.7613} & \textbf{24.726} \\
\hline
Corn    & ActiveInitSplat  & \textbf{0.4310} & \textbf{0.7920} & \textbf{21.649} \\
        & 3DGS-FVS         & 0.5155 & 0.7560 & 19.450 \\
\bottomrule
\end{tabular}}
\end{table}

\vspace{0.1cm}

\noindent \textit{\textbf{Q2.} How does ActiveInitSplat perform compared to alternative existing active coverage-based image selection strategies?}

\vspace{0.1cm}

\noindent To further substantiate the advantages of the proposed active image selection strategy in ActiveInitSplat, beyond comparisons with passive baselines, we additionally compare it against an alternative active approach grounded in the farthest view sampling (FVS) principle, which aims to select views that maximize diversity across view space and scene content; see e.g. \cite{xiao2024nerf}. It is worth noting that while FVS has been utilized in various settings, including NeRFs, it has not yet been used for GS in the context of 3D scene rendering; yet we integrate FVS into GS as a baseline to underscore the effectiveness of ActiveInitSplat. Table \ref{tab:fvs_comparison} summarizes the rendering performance of the competing approaches, where it can be clearly seen that ActiveInitSplat achieves superior rendering performance on almost all datasets, highlighting its advantages over the active FVS-based alternative.

\vspace{0.1cm}

\noindent \textit{\textbf{Q3.} How does ActiveInitSplat perform when (i) alternative methods (other than COLMAP) are used to generate $\mathcal{P}_C(\cdot)$ in the optimization process, and (ii) only a limited number of training images are acquired during the active selection process?}

\vspace{0.1cm}

\noindent Here, the goal is to show the ability of ActiveInitSplat in efficiently integrating alternative COLMAP-free approaches for PC generation within the active view selection process, while accounting for practical  settings where only a limited number of training images can be acquired. To that end, we leverage the pre-trained MASt3R model \cite{leroy2024grounding} for 3D point cloud generation, in a similar way as in the sparse-view based GS model in \cite{fan2024instantsplat}. Specifically, the $r_q$ function at each iteration incorporates the MASt3R-generated $\mathcal{P}_C(\cdot)$ and we set the total number of iterations to be 12 to accommodate sparse-view settings. For the experimental evaluation again we consider five different test sets of 20 images and report the average performance. Table \ref{tab:quant_mast3r} presents the LPIPS, SSIM, and PSNR rendering metrics of ActiveInitSplat compared to passive baselines on three indicative real-world datasets for 3DGS utilizing MASt3R-based Gaussian initialization. The results clearly indicate that ActiveInitSplat outperforms passive counterparts, highlighting its effectiveness in integrating alternative, efficient COLMAP-free point cloud generation techniques into the active view selection process. This demonstrates that ActiveInitSplat not only alleviates the computational burden of passive GS with densely collected images but also enhances GS models under sparse-view conditions. It is also worth noting that while MASt3R produces a significantly denser point cloud than COLMAP for GS initialization with only a few images, the sparsity of views may still degrade GS rendering performance. In contrast, a larger training set from COLMAP with \textit{diverse} viewpoints, as captured by our ActiveInitSplat, can lead to improved results.

\vspace{0.1cm}

\noindent \textbf{Runtime comparison.} To demonstrate the time efficiency of ActiveInitSplat, Table \ref{tab:running_time} reports the total runtime for collecting and processing $T$ images for 3D point cloud generation using MASt3R, which is subsequently employed for GS initialization. As shown in Tables \ref{tab:quant_mast3r} and \ref{tab:running_time}, when using $T=12$, ActiveInitSplat attains higher rendering quality than the passive baselines with only a negligible increase in runtime. This implies that more images must be passively collected to meet the rendering performance of ActiveInitSplat. However, increasing the number of images can markedly increase the overall processing cost, as reflected in the runtime results for passive selection with $T = 20$ and $T = 30$ in Table \ref{tab:running_time}.

\vspace{0.1cm}

\noindent \textbf{Remark 2.} It is worth noting that even for the more time-intensive SfM process, its computational overhead can be significantly reduced through incremental updates as new images are acquired online in the active selection process of ActiveInitSplat, instead of re-running the process from scratch. 

\vspace{0.1cm}

\noindent \textit{\textbf{Q4.} How does ActiveInitSplat perform when alternative GS architectures are used?}

\vspace{0.1cm}

\noindent Here, our aim is to demonstrate the effectiveness of ActiveInitSplat while remaining agnostic to the underlying GS architecture. In addition to the traditional 3DGS model in \cite{kerbl20233d} that relies on the SfM process for initialization, in \textit{Q2} we have also tested ActiveInitSplat on a different GS model that alternatively uses the MASt3R pre-trained model in a similar way as in \cite{fan2024instantsplat}. To further support our claim, we next integrate the proposed ActiveInitSplat into alternative GS architectures, including CoR-GS \cite{zhang2024cor}, and Compact-3DGS \cite{lee2024compact}. Table \ref{tab:corcompact} demonstrates the rendering performance of ActiveInitSplat compared to passive alternatives when using CoR-GS and Compact-3DGS. It is evident that ActiveInitSplat has superior performance compared to the passive counterparts in both CoR-GS and Compact-3DGS, underscoring its versatility and architecture-agnostic advantages across different GS frameworks.

\vspace{0.1cm}





\section{Conclusions}


The present work introduces `ActiveInitSplat', a novel approach for actively selecting camera views to capture informative training images, assisting both the initialization and training of GS models. In contrast to existing GS methods that rely on passively pre-selected training images, the proposed framework dynamically acquires 2D images, optimizing density and occupancy-based criteria of the constructed PC from these images. By actively selecting images from diverse viewpoints, ActiveInitSplat ensures comprehensive scene coverage and properly initializes Gaussian functions, conforming them to the ground-truth 3D structure. Numerical tests conducted on both simulated and well-established benchmark datasets highlight the significant advantages of ActiveInitSplat over conventional passive selection approaches commonly used in existing GS methods. 

\noindent \textbf{Limitations.} A basic limitation of the current work is its adaptivity to dynamic changes in the targeted scenes, since it is designed for static environments. Extending ActiveInitSplat to further account for dynamic 3D scenes belongs to our future research agenda.

{\small
\bibliographystyle{ieeenat_fullname}
\bibliography{main}
}

\end{document}